\documentclass[runningheads]{llncs}
\usepackage[T1]{fontenc}
\usepackage{graphicx}
\usepackage{cite}

\usepackage{hyperref}
\usepackage{color}

\usepackage{color}
\usepackage{epstopdf}

\begin{document}

\title{Automatic Piecewise Linear Regression for Predicting Student Learning Satisfaction}

\titlerunning{APLR for Predicting Student Learning Satisfaction}

\author{%
    Haemin Choi\inst{1}\orcidID{0009-0004-0639-1038} \and \\
    Gayathri Nadarajan\inst{1}\orcidID{0000-0001-8601-0346}
}

\authorrunning{H. Choi \and G. Nadarajan}

\institute{%
    Department of Data Science, School of Covergence, \\
    College of Computing and Informatics,\\
    Sungkyunkwan University, South Korea\\
    \email{chm1009@g.skku.edu, gaya@g.skku.edu}
}

\maketitle        

\begin{abstract}
Although student learning satisfaction has been widely studied, modern techniques such as interpretable machine learning and neural networks have not been sufficiently explored. This study demonstrates that a recent model that combines boosting with interpretability, automatic piecewise linear regression (APLR), offers the best fit for predicting learning satisfaction among several state-of-the-art approaches.
Through the analysis of APLR's numerical and visual interpretations, students' time management and concentration abilities, perceived helpfulness to classmates, and participation in offline courses have the most significant positive impact on learning satisfaction.
Surprisingly, involvement in creative activities did not positively affect learning satisfaction.
Moreover, the contributing factors can be interpreted on an individual level, allowing educators to customize instructions according to student profiles.

\end{abstract}
\section{Introduction}
\label{intro}

Student learning satisfaction has been a key concern for educators and learners alike~\cite{long1985contradictory, topala2014learning}.
Understanding the determinants of student learning satisfaction would enable educators and institutions to tailor better instruction methods to enhance the overall learning experience. In this study, we focused on mining the factors affecting student learning satisfaction during the COVID-19 pandemic, including demographics, learning methods, perceived performance, self-efficacy, motivation, engagement, emotional state, stress coping mechanisms and learning environment.
We conducted a cross-sectional study on 302 students from 
Sungkyunkwan University, South Korea, 
of diverse study majors, course types 
and learning methods, 
after they were exposed to online learning for two years and could better gauge their preferences. We employed a recent interpretable machine learning method, APLR, that provides visual explanations on the model’s decisions. The implementation code and data are available at \href{https://github.com/jaime-choi/Automatic-Piecewise-Linear-Regression-for-Predicting-Student-Learning-Satisfaction}{github.com/jaime-choi/APLR-for-Predicting-Student-Learning-Satisfaction}.

The key contributions of this work are twofold:
1) APLR outperforms representative bagging and boosted trees, an interpretable additive model,
as well as 
a transformer-based deep learning model to predict learning satisfaction in four out of five metrics, and; 2) The global and local interpretations of APLR provide valuable insights into the factors influencing learning satisfaction on the overall group and on individual students, paving the way for personalized learning.

This paper is structured as follows. Section~\ref{rw} outlines existing efforts and gaps in education technology and machine learning for mining learning satisfaction. Section~\ref{met} provides the methodology and experimental setup for APLR and its competitors. The results of the models and APLR's interpretations are detailed in Section~\ref{res}. Section~\ref{conc} concludes and discusses future directions.

\section{Related Works}
\label{rw}
We adopt the definition of learning satisfaction outlined by Chang \& Chang~\cite{chang2012effect} as the “perceived level of fulfillment connected to the individual’s desire to learn, caused by the learning motivation”. We are particularly interested in learning during the pandemic period whereby different learning methods, emotional states, coping mechanisms and learning environments could have more pronounced impact on their learning satisfaction, beyond what traditional studies have done when learning was not hindered by as many challenging factors.

\subsection{Studies on Student Learning Satisfaction}

Several notable studies found a positive correlation between students’ self-efficacy and online learning satisfaction~\cite{aldhahi2022exploring, alqarashi2019, chu2010multi, kirmizi2015influence}. Self-efficacy is students’ beliefs in their capabilities to perform learning tasks. Student engagement is another factor that affects online learning satisfaction in a positive way~\cite{she2021online,sinval2021university}. Some studies looked more specifically at interactions
between learners~\cite{kurucay2017examining,skinner2008engagement}, and learners and educators~\cite{baber2020determinants,costley2016effects} and found positive influences on satisfaction. Our study looked at interaction between learners and also feedback from instructors.

Other studies have found multiple factors affecting learning satisfaction, to name a few: Eom \& Ashill~\cite{eom2016determinants} showed that engagement and course design significantly affect students' satisfaction and learning outcomes, Ikhsan et al.~\cite{ikhsan2019determinants} found that engagement, technical support and student motivation affect the learning outcomes and satisfaction, Huang~\cite{huang2021using} discovered that perceived usefulness and perceived ease of use have a positive impact on learning satisfaction in blended learning, and Ren et al.~\cite{ren2024factors} demonstrated that the teaching environment and quality of teachers’ online learning are important factors for blended learning.

A bulk of these studies~\cite{baber2020determinants,eom2016determinants,ikhsan2019determinants,huang2021using,ren2024factors,she2021online} incorporated structural equation modeling (SEM)~\cite{ullman2012structural}, an integrated framework of several multivariate techniques such as latent variables, path analysis, regression, measurement theory and simultaneous equations within one model, to mine education data. Others have used statistical hypothesis testing as a major component of their analysis. It should be noted that the current literature on education data mining rarely considered emotional and learning environment factors. A recent study by Han et al.~\cite{han2024exploring} included these considerations and used an Explainable Boosting Machine (EBM) model~\cite{lou2013accurate} to mine student learning satisfaction during the pandemic. EBM belongs to the class of intrepretable machine learning models (see next section). They modeled the learning satisfaction target as a loose combination of two features only. Building upon this, we altered the target to include a wider definition of learning satisfaction and utilized a more recent interpretable model that outperforms EBM and other competing models. 

\subsection{Interpretable ML and Neural Networks}
\label{ainn}

The field of trustworthy Artificial Intelligence has garnered a fair amount of attention recently~\cite{gilpin2018explaining,roscher2020explainable}. Machine learning (ML) models are known to achieve high performance but the decision-making process of many, such as deep neural networks, are not known to humans (not interpretable or black boxes). 
It is widely known that there exists a trade-off between performance and interpretability. While methods like SEM could provide a good understanding of relationships between features, it requires specific assumptions about the data and cannot make predictions on new data. ML models are more flexible, can be used to make predictions on new data and tend to be more accurate. Moreover, interpretable ML models such as decision trees, EBM and APLR are inherently human-understandable, while black box models such as random forests and deep neural networks can be made human-understandable with the addition of explainable modules with a slight reduction in accuracy. Hence, we focus on ML models, specifically those comparable to APLR, for our work. 

Random forest (RF)~\cite{breiman2001random} is 
a learning method that takes the combination of multiple decision trees to reach a final result. 
In each iteration of a RF, a random sample of data in a training set is selected with replacement (boostrapped dataset) and fitted to a tree using a random subset of the input features.  This random selection of dataset and tree fitting to a subset of features is repeated to produce a variety of trees, making it more effective than building just one tree. For classification, it takes the class with the majority votes (aggregation) as the final prediction for a new data instance.

Light Gradient Boosting Machine (LightGBM)~\cite{ke2017lightgbm} is also a tree-based learning method but within a gradient boosting framework. A boosting algorithm trains models sequentially, with each model learning from the errors of the previous one, while a bagging algorithm such as RF trains models on different subsets of data in parallel and aggregates them. LightGBM has been dominant among the boosting algorithms in terms of speed. For example, when splitting a tree, LightGBM works on discrete bins of a histogram instead of continuous values, making it highly efficient. Another feature of LightGBM is exclusive feature bundling, which reduces the number of features by merging features that are sparse. Finally, when reducing the loss in a model via gradient boosting, LightGBM only keeps the instances with large gradients (that will contribute more to the information gain) and randomly drops the instances with small gradients, using a smaller dataset overall. LightGBM has been shown to outperform the slower and yet powerful XGBoost algorithm~\cite{chen2016xgboost} in some instances~\cite{mic2017,turkmen2024}.

Explainable Boosting Machine (EBM)~\cite{lou2013accurate} is an interpretable model which is an augmentation of a generalized additive model (GAM)~\cite{hastie1987generalized}. A GAM fits one or more arbitrary functions into a generalized linear model, i.e. it finds nonlinear relationships between the target and inputs. The target is expressed as a combination of arbitrary functions of its inputs. EBM is an improved version of GAM models in that it takes into account both the relationships of each single (univariate) input with the target (like GAM) and 
the relationships of {\em pairwise interaction} of two inputs with the target. Therefore, it can interpret how each sole feature affects the target variable as well as how the interactions between two inputs influence the target.

TabNet~\cite{arik2021tabnet} is a deep learning technique for tabular data that serves as an alternative to ensemble and tree methods. It is a transformer-based model that uses sequential attention to choose which features to reason from at each decision step. 
For each step, a feature selection mask provides interpretable information about the model’s functionality, and the masks can be aggregated to obtain global feature important attribution. 
While providing interpretability, it could be prone to overfitting like other neural networks. However, we have considered to include TabNet along with the other methods in this section.

Automatic piecewise linear regression (APLR)~\cite{von2024automatic} leverages gradient boosting which utilizes an ensemble of weak learners to produce a good estimate and multivariate adaptive regression splines (MARS)~\cite{friedman1991multivariate}, which is interpretable and considers interactions among variables. It uses componentwise gradient boosting~\cite{buhlmann2003boosting} in which one simple base learner is fitted for each predictor, and the one that helps the most to minimise the loss function is kept at each step. The inner workings of APLR's boosting method can be found in Section~\ref{aplr}.
APLR is able to compete with boosted trees and RF on predictiveness for regression, however, at the time of writing, it has not been tested rigorously on classification tasks. Compared to EBM, APLR splits the data into segments and fits a linear model to each segment, while EBM is additive, i.e. the target variable is the sum of smooth (non-linear) functions of individual features. APLR is also computationally more efficient than EBM and comparatively easier to use than RF and boosted trees~\cite{von2024automatic}.

Despite the availability of a plethora of interpretable and explainable AI models, there has been a lack of effort in applying these methods in education data mining. Our work aims to address this gap with the application of APLR for 
predicting student learning satisfaction and their determinants. 

\vspace{-0.15cm}

\section{Methods}
\label{met}
\subsection{Survey Dataset}
To capture the multifaceted nature of students’ learning satisfaction during the pandemic, an online survey containing questions about overall learning experiences was conducted with 302 students from Sungkyunkwan University after four semesters of online learning (late 2021-late 2022). The survey included questions on students’ emotional state, stress management, and online learning environment. The survey participants consisted of full-time and exchange students in South Korea, which made up 88\% and 12\% of the total respectively. The gender and majors of the participants were mixed and balanced. A vast majority of the courses (76.82\%) were conducted live online, via pre-recorded lectures or flipped learning, with the rest conducted in-class (offline).
The responses ranged from 5-point Likert-type values (“Strongly disagree” to “Strongly agree”), to binary (“Yes”/“No”), to one-of-$N$ answers for the factuality of the information. The students' majors were grouped into Science, Technology, Engineering and Medicine (STEM) (41.4\%), Humanities and Social Sciences (HSS) (40.6\%), or hybrid categories (18\%), and questions were encoded into simpler column names for easier processing and analysis. Likert values of responses were encoded to numerical values for classification modeling and factor analysis\footnote{The entire survey is available at~\href{https://forms.gle/3bEMbszuUDW3MnQx5}{https://forms.gle/3bEMbszuUDW3MnQx5}}.

\subsection{Task Design}
 We designed a binary classification task for the target variable {\em learning satisfaction}, which was constructed using seven features, which were adapted from twenty four questions developed by Bolliger \& Halupa~\cite{bolliger2012student} on learning satisfaction. For each positive feature in Table~\ref{tab1}---those not ending with {\em “(Neg)”}---response values “Strongly disagree” and “Disagree” were encoded as -1, “Neutral” as 0, and “Agree” and “Strongly agree” as 1 while the negative features {\em m\_feedback} and {\em emo\_miss} were encoded the opposite way since a disagreement would imply satisfaction. The satisfaction score was constructed by summing the encoded numerical values of the seven features, which could range from -7 to 7. 
Samples with satisfaction scores totaling 4 or more were coded as positive (1, “Satisfied”), and the rest as negative (0, “Not satisfied”). Compared to the task modeled by Han et al.~\cite{han2024exploring} which loosely defined the target using an OR combination of $m\_valuable$ and $m\_taskPerformance$, we expanded the target to encompass a broader definition of learning satisfaction. The task used 47 input features, or predictors with response values ranging from -2 (“Strongly disagree”) to 2 (“Strongly agree”).

\vspace{-0.15cm}

\begin{table}
\caption{Features that constitute the target variable, {\em learning satisfaction}. A combination of four or more positive values would imply a positive target value. Negative features, ending with encoding {\em (Neg)}, are flipped before being aggregated.}\label{tab1}
\setlength{\tabcolsep}{0.5em}
\begin{tabular}{p{8cm} p{3cm}}
\hline
Survey Question &  Encoding \\
\hline
The learning method is suitable for this course. &  {\itshape m\_suitable} \\
I feel comfortable with the way this course is conducted.  &  {\itshape m\_comfortable} \\
I am sometimes frustrated because I cannot get instant feedback. & {\itshape m\_feedback (Neg)} \\
I believe the things we studied in this course could be of & {\itshape m\_valuable}\\ some value to me. &  \\
I would be willing to take other courses with the same & {\itshape m\_sameMethod} \\ 
learning method again. & \\
I am satisfied with my performance at the tasks given in the lessons. & {\itshape m\_taskPerformance} \\
I feel I have been missing out on proper learning. & {\itshape emo\_miss (Neg)} \\
\hline
\end{tabular}
\end{table}

\vspace{-0.15cm}

\subsection{Data Distribution}
After the construction of the target variable, the survey data was split into training and test sets in a 0.8:0.2 ratio with 241 and 61 samples, respectively. Since the training data was imbalanced with more positive samples, the Synthetic Minority Over-sampling Technique (SMOTE)~\cite{chawla2002smote} was applied.

\vspace{-0.15cm}

\subsection{Automatic Piecewise Linear Regression (APLR)}
\label{aplr}

As mentioned in Section~\ref{rw}, APLR is a gradient-boosting method with the application of base functions inspired by MARS. Being the building block of boosting, a base function captures the effect of the predictors on the response through the negative gradient. Along with the simple linear effect, APLR further uses specific basis functions that capture non-linearity and interactions through local effects.

The componentwise boosting step for each $m=1$ to $M$ in the APLR fitting procedure for regression is as follows (\hspace{1sp}\cite{von2024automatic}, Sec. 3.2.2): 
\begin{enumerate}
    \item Compute the negative gradient $u_m$ using the squared error loss function.
    \begin{equation}\label{eq1}
        u_{m} = y - \hat{f}_{m-1}(C_{m-1})
    \end{equation}
    Here, $\hat{f}_{m-1}(C_{m-1})$ is the estimated response, and $C_{m-1}$ is the set of non-intercept terms in the model at the previous boosting step, $m-1$.
    \item Initialize $C_m$ = $C_{m-1}$. The intercept is updated using the (weighted) mean of $u_m$ multiplied by the learning rate $v 
    \in(0,1]$. Afterwards, the negative gradient is recomputed.
    \item For each term $e_j$ in $E$ (the eligibility of terms), find the APLR basis function $h_m(u_m, e_j)$ that fits best to $u_m$ by having the lowest loss.
    \item Select the term with the lowest loss $h_m(u_m, e_*)$ as a candidate to enter $C_m$. In this step, interaction terms are considered together.
    \item Update the regression coefficient for the term from steps 3 and 4 that reduced the loss the most. This term is added to $C_m$ unless it is already in $C_m$.
\end{enumerate}

\noindent In step 5, the weighted linear regression coefficient $\beta$ of an APLR basis function $f(x)$ is estimated as shown in Equation~\ref{eq2}:
\begin{equation}\label{eq2}
\beta = v \cdot \frac{\sum_{i=1}^{n_{eff}} f(x_{i}) \cdot w_{i} \cdot u_{m,i}}{\sum_{i=1}^{n_{eff}} f(x_{i})^{2} \cdot w_{i}}
\end{equation}

$n_{eff}$ is the number of effective observations and $w$ is the sample weight provided by the user. If $w$ is not provided, $\beta$ is estimated without it. The regression coefficient for each term can be used to interpret the effect and importance of the term, further providing users with an understanding of how individual features and interactions between two features affect the prediction of the target. 

For the APLR application for binary classification, the binomial negative log-likelihood is set as the loss function and the logit function as the link function. The classifier is fit to a logit model for each response category and the class with the highest predicted probability is chosen. In our case, two logit models (positive and negative) are fit and the interpretations are given in Section~\ref{res}.

\vspace{-0.15cm}

\subsection{Hyperparameter Tuning}
\label{hyper}
We tuned APLR and four other models before comparing their performance on our task. See Section~\ref{ainn} for a more thorough treatment of these models. The hyperparameters of Random Forest, LightGBM, and APLR were tuned as prescribed by the authors of APLR. 
For EBM, we used the default hyperparameters that were employed by Han et al.~\cite{han2024exploring} that should perform well on most problems. For TabNet, a learning rate of 0.02 with decay was set as suggested by the authors~\cite{arik2021tabnet}, and a smaller batch size was used to suit our dataset size. 
We outline the tuning procedure for the first three models below and provide the results in Table~\ref{tab2}.

\vspace{-0.15cm}
\begin{table}[!h]
\centering
\caption{Optimal hyperparameter values for APLR and four other competing models. TabNet. A fixed seed value of 42 was used in all settings to prevent inconsistent results.}\label{tab2}
\setlength{\tabcolsep}{0.5em}

\begin{tabular}{|l|l|p{5.5cm}|}
\hline
Method &  Hyperparameter Range & Optimal Value \\
\hline
Random Forest & \{0.125, 0.25, 0.5, 0.75, 1\} & $max\_features$ = 0.25 \\
 & \{1, 20, 50, 100, 500\} & $min\_samples\_leaf$ = 1 \\
 & \{100, 300, 500\} & $n\_estimator$ = 500 \\
\hline
LightGBM & [1, 30000] & $num\_estimators$ = 2910 \\
 &  [2, 128] & $num\_leaves$ = 4 \\
\hline
APLR & \{0, 1, 2, 100\} & $max\_interaction\_level$ = 1  \\
 & \{20, 100, 500\} & $min\_observations\_in\_split$ = 20 \\
\hline
EBM & & $max\_leaves$ = 3 \\
& Not Applicable & $smoothing\_rounds$ = 75 \\
& (Default) & $learning\_rate$ = 0.015 \\
& & $interactions$ = 0.9 \\
\hline
TabNet & & $lr$ (optimizer\_params) = 2e-2 \\
& Not Applicable & $step\_size$ = 50\\
& (Default) & $gamma$ = 0.9 (scheduler\_params)\\
& & $max\_epochs$ = 200 \\
& & $batch\_size$ = 32 (smaller than default)\\
& & $virtual\_batch\_size$=16 \\
\hline
\end{tabular}
\end{table}

\vspace{-0.15cm}

\subsubsection{Random Forest.} The hyperparameters considered were $max\_features$ (maximum number of features to split a node), $min\_samples\_leaf$ (minimum number of observations required in a node), and $n\_estimator$ (number of trees). All hyperparameters were tuned using grid search across five-fold cross-validation.

\vspace{-0.15cm}

\subsubsection{LightGBM.} The hyperparameters considered were $n\_estimators$ (number of boosting steps), $num\_leaves$ (maximum number of leaves in each tree), and $v$ (learning rate). $n\_estimators$ and $num\_leaves$ were tuned by using the Bayesian probabilistic model-based approach provided in the \texttt{Optuna} package for Python~\cite{optuna}. 
$v$ was set to 0.1 and $n\_trials$ (unique combination of hyperparameters) was held at 100 to allow 100 unique combination trials.

\vspace{-0.15cm}

\subsubsection{APLR.} The hyperparameters considered were $max\_interaction\_level$ (maximum allowed depth of interaction terms) and $min\_observations\_in\_split$ (minimum effective number of observations that a term must rely on). They were tuned using the built-in five-fold cross-validation grid search, \texttt{APLRTuner}~\cite{von2024automatic}. $M$ (boosting steps) was held constant at 3000 and $v$ (learning rate) was set to 0.5 as suggested by the authors of APLR~\cite{von2024automatic}. 

\vspace{-0.15cm}

\section{Results}\label{res}

\subsection{Performance Comparison}
\label{results}
To evaluate the performance of the models implemented for our binary classification task, the metrics accuracy, F1, precision, recall, and AUC (Area Under the Receiver-operating Characteristic Curve) were calculated.

As shown in Table~\ref{tab3}, APLR outperformed other competing models in almost all metrics and comes a close second behind Random Forest in AUC score. Overall, APLR achieves better predictiveness than existing representative bagging, boosting and deep learning algorithms.
Furthermore, APLR outperformed EBM, an interpretable competitor algorithm while capturing non-linearity and interaction relationships. The structured and small-scale nature of our survey dataset contributed to APLR outperforming TabNet. Parametric deep  neural networks such as TabNet often struggle to generalize effectively on small-scale datasets, due to their capacity to model complex and high-order feature interactions. In contrast, the structured and ordinal nature of our dataset is more appropriately modeled using an additive and segmented approach. From a pedagogical perspective, while there is not one model that fits all types of datasets, these results suggest that APLR would be a fitting choice for structured, small-scale data.

\vspace{-0.15cm}

\begin{table}
\centering
\caption{Performance results of APLR and four competing models for student learning satisfaction classification task.}\label{tab3}
\setlength{\tabcolsep}{0.6em}
\def\arraystretch{1.3}
\begin{tabular}{l l l l l l}
\hline
 &  Accuracy & F1 & Precision & Recall & AUC \\
\hline
APLR &  {\bf 0.885} & {\bf 0.909} & {\bf 0.921} & {\bf 0.897} & 0.926 \\
\hline
Random Forest & 0.820 & 0.853 & 0.889 & 0.820 & {\bf 0.947} \\
\hline
LightGBM &  0.803 & 0.846 & 0.846 & 0.846 & 0.889 \\
\hline
EBM & 0.820 & 0.853 & 0.889 & 0.821 & 0.918 \\
\hline
TabNet & 0.836 & 0.872 & 0.872 & 0.872 & 0.818 \\
\hline
\end{tabular}
\end{table}

\vspace{-0.15cm}

\subsection{APLR Terms and Coefficients}
\label{terms}

Table~\ref{tab4} presents the first ten terms added to the logit model for the positive class (“Satisfied”), other than an intercept, in our classification task. A term can be made up of a single feature (e.g. P0) or an interaction of features (e.g. P7). Here, the intercept is -1.832 and the first seven terms added to the model are the predictors P0 to P6. 
These are basis functions without interactions that represent simple linear effects of each predictor. For example, there is an expected increase of 0.447 in the log odds of the response ({\it learning satisfaction}) for each unit increase of $m\_timeManage$ (P0) and a decrease of 0.15 for each unit increase of $cop\_creative$ (P6). See Table~\ref{tab5} for the complete definition of these encodings.

The term P7 represents an APLR basis function with interactions between the features $m\_ta$ and $m\_helpful$.
It only contributes to the prediction when $m\_ta$ is less than 0 and $m\_helpful$ is non-zero, in which there is an expected increase of 0.374 in the log odds of the response for each unit increase of $m\_ta$. The coefficient values are the inverse (same value with the opposite sign) for the logit model for the negative class (“Not satisfied”). APLR provides an added functionality to interpret a model by estimating global feature importance and local feature contributions, which will be discussed next.

\begin{table}
\centering
\caption{The first ten predictors added to the logit model for the positive class (“Satisfied”), intercept, and their coefficient values.} \label{tab4}
\setlength{\tabcolsep}{0.5em}
\begin{tabular}{l|l|l}
\hline
Interaction Level & Predictor & Coefficient \\
\hline
$null$ & Intercept & -1.832 \\
\hline
0 & P0: $m\_timeManage$ & 0.447 \\
\hline
0 & P1: $m\_concentrate$ &  0.416 \\
\hline
0 & P2: $m\_helpful$ & 0.34 \\
\hline
0 & P3: $mode\_Offline$ & 0.73 \\
\hline
0 & P4: $m\_boring$ & 0.222 \\
\hline
0 & P5: $emo\_isolated$ & 0.193 \\
\hline
0 & P6: $cop\_creative$ & -0.15 \\
\hline
1 & P7: min($m\_ta$-0,0) * I($m\_helpful$!=0) & 0.374 \\
\hline
1 & P8: $m\_boring$ * I($m\_timeManage$!=0) & 0.124 \\
\hline
1 & P9: $m\_timeManage$ * I(max($m\_boring$-0,0)!=0) & 0.147 \\
\hline
\end{tabular}
\end{table}

\vspace{-0.15cm}

\subsection{Global and Local Interpretations}
\label{interp}


\begin{figure}[!h]
    \centering
    \includegraphics[width=\textwidth]{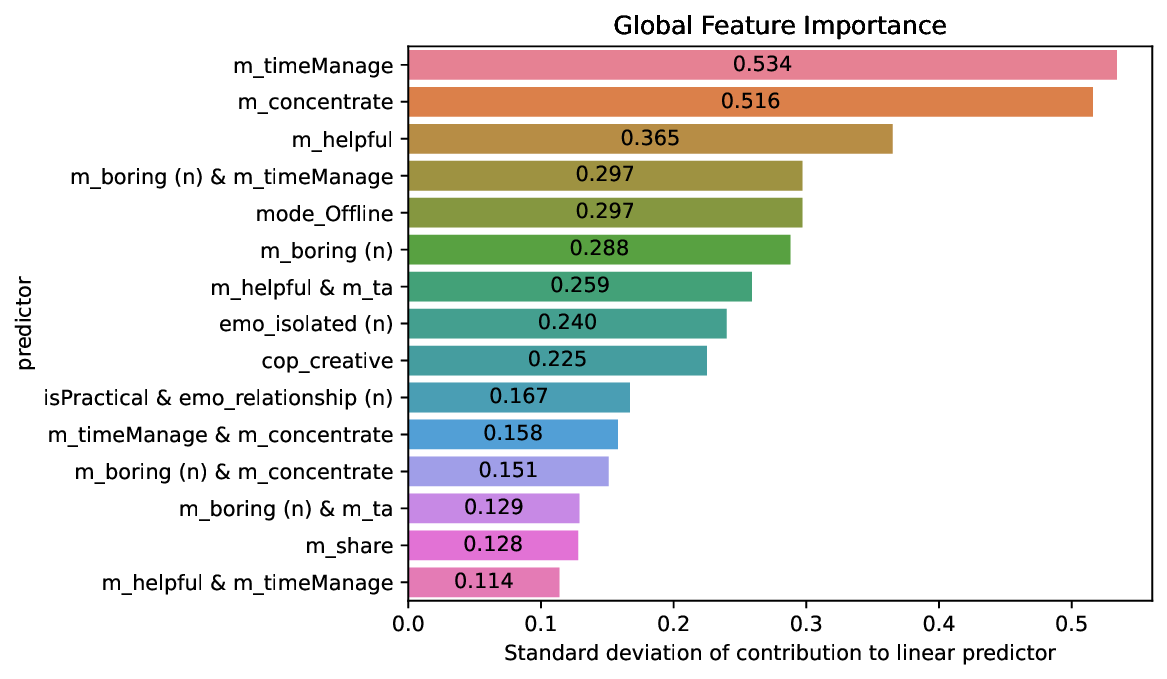}
    \caption{The top fifteen contributing predictors and their global feature importance for classifying learning satisfaction using APLR.}
    \label{global-interp}
\end{figure}

Fig.~\ref{global-interp} shows the top fifteen global features estimated by the APLR classifier. 
To calculate a feature's importance, the contributions of all terms having that feature as the main predictor are used to calculate the standard deviation in the training data. 
The global feature importance values in Fig.~\ref{global-interp} are the average across our two logit models and can be interpreted as the significance of the feature in predicting learning satisfaction.
Table~\ref{tab5} provides the encoded feature names and their corresponding survey questions.

For our task, $m\_timeManage$, $m\_concentrate$, and $m\_helpful$ were found to have the strongest influence on predicting students’ learning satisfaction, with feature importance values 0.534, 0.516, and 0.365 respectively. Since the coefficient of $m\_timeManage$ (P0) in Table~\ref{tab4} is positive (0.447), students who believe they can effectively manage their time make the most significant contribution to the prediction of the positive class (“Satisfied”). In contrast, those who struggle with time management contribute the most to the prediction of the negative class (“Not satisfied”). 
Similar patterns emerge for students who can concentrate well while studying ($m\_concentrate$), perceive themselves as helpful to classmates ($m\_helpful$), and who take offline courses ($mode\_Offline$). 
An increase in any of these features generally increases learning satisfaction. Two negative features, $m\_boring (n)$ and $emo\_isolated (n)$ have feature importance values 0.288 and 0.240 while having positive coefficient values in Table~\ref{tab4}, 0.222 and 0.193 respectively. Since they were encoded in the opposite manner, we can intuitively interpret this as students who do not find the course boring or do not feel isolated are still satisfied with their learning, though the impact of these features are not as significant as that of $m\_timeManage$ or $m\_concentrate$. 

The fourth-ranked predictor $m\_boring (n)$ \& $m\_timeManage$, is made up of two interacting features. 
We can loosely say that students who do not find the course boring and can manage their time well are satisfied with their learning. In  Table~\ref{tab4}, we can see two terms (P8 and P9) containing the interaction between these two features, both having positive coefficients to support this. Conversely, $cop\_creative$ with global feature importance value 0.225 has a negative coefficient (-0.15) in Table~\ref{tab4}. This can be interpreted as students who engage in creative activities tend to be slightly unsatisfied with their learning.

\begin{figure}
    \centering
    \includegraphics[width=\textwidth]{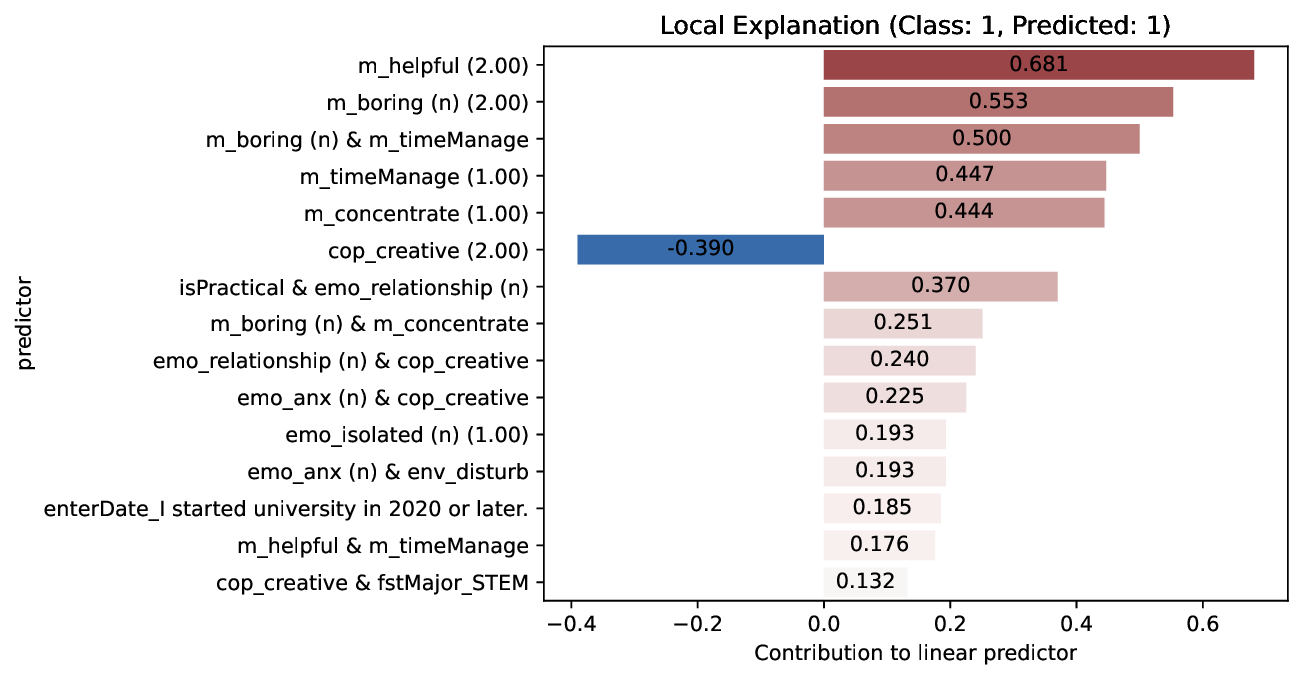}
    \caption{APLR's local explanation when making a prediction for a sample from the positive class. All bars on the right indicate positive influence, with longer bars showing stronger influence. The bar on the left indicates negative influence.}
    \label{local-interp-pos}
\end{figure}

\begin{figure}
    \centering
    \includegraphics[width=\textwidth]{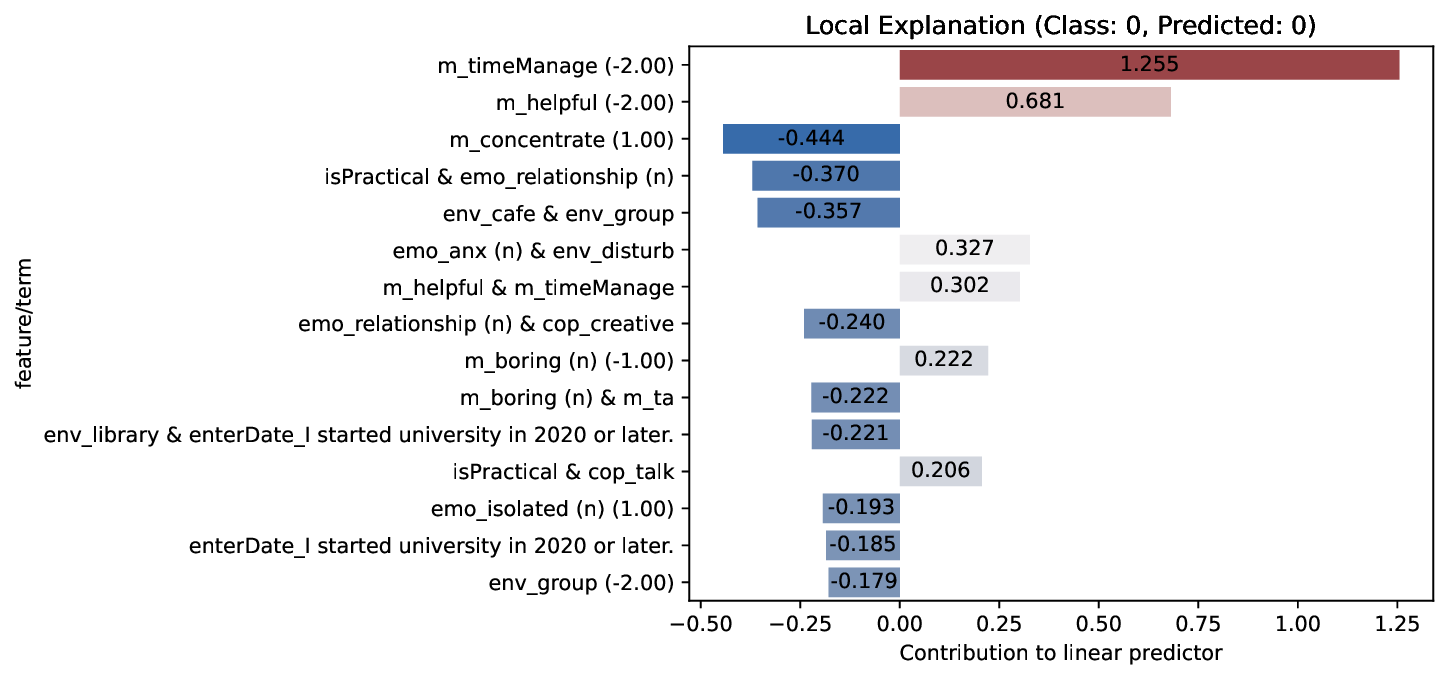}
    \caption{APLR's local explanation when making a prediction for a sample from the negative class. All bars on the right indicate positive influence towards predicting the negative class, with longer bars showing stronger influence.}
    \label{local-interp-neg}
\end{figure}

Fig.~\ref{local-interp-pos} presents the local feature contributions to the logit model for a sample from the positive class (“Satisfied”). Here, we visualize the contributions of individual features when classifying a single sample from the (unseen) test dataset of a student. 
The features $m\_helpful$ and $m\_boring (n)$ are the most influential in making the prediction for this sample, with contributions of 0.681 and 0.553 respectively. They both have the survey response value 2.00 (\textit{Strongly Agree}) for this sample. The fact that this student can be helpful to her/his classmates and not find the course boring are the biggest contributors to their learning satisfaction.
In Fig.~\ref{local-interp-pos}, $m\_timeManage$ and $m\_concentrate$ 
with value (1.00, \textit{Agree})
also show high contribution values, 0.447 and 0.444.
This highlights a concrete example in which perceived helpfulness, not feeling boredom, good time management and concentration skills lead to learning satisfaction.
In contrast, $cop\_creative$ shows a negative contribution (-0.390), given that its value is positive for this student (2.00, \textit{Strongly Agree}). 
A possible interpretation of this is that the student is involved in creative activities such as art or music but this involvement does not contribute towards their (positive) learning satisfaction.

Fig.~\ref{local-interp-neg} presents the local feature contributions to the logit model for a sample from the negative class (“Not Satisfied”). The features $m\_timeManage$ and $m\_helpful$ are the most influential in predicting the class for this sample, with contributions of 1.255 and 0.681. Given that their values are negative (-2.00, \textit{Strongly Disagree}), we can infer that this student's poor perceived time management skills and apparent inability to help classmates appear to have significantly contributed to the prediction of the negative class. In contrast, the value for $m\_concentrate$ is positive (1.00, \textit{Agree}) while showing a negative contribution (-0.444). 
This indicates that this student's concentration capability is inversely proportional to her/his dissatisfaction with learning. 

To summarize, even though the local feature contribution rank shows a similar trend with global feature importance, the details can differ by data instances, as it provides an interpretation of a specific sample (student). This can help educators customize their instructions for students with different learning profiles.

\begin{table}[!h]
\caption{Selected survey questions and their feature encodings. Responses are 5-point Likert-type values, except for three, specified in brackets at the end of the questions.}\label{tab5}

\setlength{\tabcolsep}{0.5em}
\begin{tabular}{p{8.6cm} p{2.9cm}}
\hline
Survey Question &  Encoding \\
\hline
The learning method is suitable for this course. & $m\_suitable$ \\
I feel comfortable with the way this course is conducted.  &  $m\_comfortable$ \\
I could manage my time properly. & {\itshape m\_timeManage} \\
I can concentrate well when studying for this course. & {\itshape m\_concentrate} \\
I think that I could be helpful to my classmates.
& {\itshape m\_helpful} \\
I find this course boring. & {\itshape m\_boring (Neg)} \\
I get support from the teaching staff when I have trouble with the course. & {\itshape m\_ta} \\
I share my opinions during class discussions.
& {\itshape m\_share} \\
I feel more isolated. & {\itshape emo\_isolated (Neg)} \\
I have lost friendships/relationships during this period. & {\itshape emo\_relationship (Neg)} \\
I often feel distressed/anxious. & 
{\itshape emo\_anx (Neg)} \\
I do creative things like art, writing, composing music, and gardening to relieve stress. & {\itshape cop\_creative} \\
I talk to people about my problems. &
{\itshape cop\_talk} \\
How is most of this course conducted? Select one. (Live Online/Pre-recorded/Offline/Flipped course) & {\itshape mode} \\
The course is practical. (Yes/No) &
{\itshape isPractical} \\
Year of entry to the university. (before 2020/2020 or later) & 
{\itshape enterDate} \\
I study in a cafe or place with some activity because I feel isolated in a quiet place. &
{\itshape env\_cafe} \\
I study in the library because I can focus better than at home. &
{\itshape env\_libaray} \\
I normally study with one or two friends during the pandemic. &
{\itshape env\_group} \\
I feel distracted at home due to people, pets, TV, etc. &
{\itshape env\_disturb} \\

\hline
\end{tabular}
\end{table}

\section{Conclusions and Future Directions}
\label{conc}
In this study, we explored various factors that influenced student learning satisfaction in South Korea during the pandemic using an interpretable ML model, APLR. We found that APLR outperformed four other highly ranked approaches in four out of five metrics. In addition, we examined the interpretability of APLR by analyzing term coefficients and feature importance. Our findings indicate that key factors such as student time management and concentration abilities, perceived helpfulness to classmates, and participation in offline courses significantly impacted learning satisfaction. Notably, agreement with these factors was associated with higher satisfaction, whereas disagreement had a negative effect.
Counterintuitively, student involvement in creative endeavors such as art and music did not contribute to their learning satisfaction. Analyzing more local explanations could potentially provide further understanding on this.

Our findings offer valuable insights for educators and institutions in designing instructional methods and educational strategies that align with students’ learning needs and satisfaction. 
Moreover, individual-level interpretability paves the way for personalized learning.

While the main aim of the work was to predict and mine the determinants of learning satisfaction for a specific dataset, it would be interesting to test APLR's performance on other similar classification datasets.
Moving forward, we aim to extend our research to investigate various aspects of students' learning experiences in the post-pandemic era, which will allow for a comparative analysis of key factors before and after the pandemic. Beyond learning satisfaction, future studies could explore other dimensions, such as students' learning motivation and perceived academic performance, to provide a more comprehensive perspective. Given the crucial role of identifying and analyzing influential factors, designing well-structured tasks and employing appropriate machine learning or deep learning methods will remain essential for future research.

\begin{credits}
\subsubsection{\ackname} The authors would like to thank Mathias von Ottenbreit for the clarifications on the interpretations of APLR.

\subsubsection{\discintname}

The authors have no competing interests to declare that are
relevant to the content of this article. 

\end{credits}
%
%
%
\bibliographystyle{splncs04} 
\bibliography{main}

\end{document}